# HIDDEN MARKOV MODEL BASED PART OF SPEECH TAGGER FOR SINHALA LANGUAGE

A.J.P.M.P. Jayaweera[1] and N.G.J. Dias[2]

[1] Virtusa (Pvt.) Ltd, No 752, Dr. Danister De Silva Mawatha, Colombo 09, Sri Lanka
[2] Department of Statistics & Computer Science, University of Kelaniaya, Kelaniya, Sri Lanka

## ABSTRACT

*In this paper we present a fundamental lexical semantics of Sinhala language and a Hidden Markov Model (HMM) based Part of Speech (POS) Tagger for Sinhala language. In any Natural Language processing task, Part of Speech is a very vital topic, which involves analysing of the construction, behaviour and the dynamics of the language, which the knowledge could utilized in computational linguistics analysis and automation applications. Though Sinhala is a morphologically rich and agglutinative language, in which words are inflected with various grammatical features, tagging is very essential for further analysis of the language. Our research is based on statistical based approach, in which the tagging process is done by computing the tag sequence probability and the word-likelihood probability from the given corpus, where the linguistic knowledge is automatically extracted from the annotated corpus. The current tagger could reach more than 90% of accuracy for known words.*

## 1. INTRODUCTION

According to figures from UNESCO (The United Nations' Educational, Scientific and Cultural Organization), there are around 6900 spoken languages are exist in this world, only 20 languages are spoken by 50% of the world population. Each of these languages are spoken by more than 50 million speakers. Most of the world population speaks Chinese Mandarin and that is spoken by around 1000 million people. Spanish, English, Hindi, Arabic, Portuguese and Russian are other top most languages spoken by the largest population in this world, and each language is spoken by 200 million speakers or more. People who speak those top most languages are spread across different geographical regions in multiple countries. Also 50% of the languages are endangered and most of them are spoken by small communities and they are always limited to a specific geographical region [1, 2, 3].

Sinhala is also one unique language that speaks only by people in Sri Lanka and more than 17 million speakers speak Sinhala as their mother tongue. We believe that Sinhala is not an endangered language yet, though speakers are limited only to a small geographical region. But we think our mother language need more attention, and need to get more provision to develop the language with latest technology trends. So our effort here is to address one pitfall that we have identified in area of computational linguistics and Natural Language Processing (NLP) related to Sinhala language.

Though research on NLP, has taken giant leap in the last two decades with the advent of efficient machine learning algorithms and the creation of large annotated corpora for various languages, only few languages in the world have the advantage of having enough lexical resources, such as English. NLP researches for Sinhala are still far behind than other South Asian languages. Further we have very limited lexical resources available for Sinhala language. Researches on NLP for Sinhala language can be pushed by creation of required lexical resources and tools. So, the attempt of this research is to develop a Part of speech Tagger for Sinhala language, which is a fundamental need for further computational linguistic analysis for our mother language.

Sinhala is a complex language, morphologically rich and agglutinative in nature, words of which are inflected with various grammatical features. Sinhala root noun (lemma) inflects for plural and singular and Sinhala verb specifies almost everything like gender, number and person markings, and represents the tense of the activity.

POS tagging is a well-studied problem in the field of NLP and one of the fundamental processing step for any language in NLP and language automation, i.e., the capability of a computer to automatically POS tag a given sentence. Throughout the history of NLP, different approaches have already been tried out to automate the task of POS tagging of languages such as English, German, Chinese and few South Asian languages such as Hindi, Tamil and Bengali.
Words are the fundamental building block of a language. Every human language, spoken, signed or written is composed of words [7]. Every area of speech and language processing, from speech recognition to machine translation, text to speech, spelling and grammar checking to language-based information retrieval on the Web, requires extensive knowledge about words that are heavily based on the lexical knowledge. In contrast to other data processing systems, language processing applications use knowledge of the language.

The basic processing step in tagging consists of assigning POS tags to every token in the text with a corresponding POS tag like noun, verb, preposition, etc., based both on its definition, as well as its context. The number of part of speech tags in a tagger may vary depending on the information one wants to capture [7].

In this paper, we present a fundamental lexical and morphological analysis of Sinhala language, theory of Hidden Markov Model and the algorithm of the implementation. Section 2 of this paper gives an idea of history and previous research on NLP and section 3 discusses previous work on Sinhala language. Section 4 and 5 give a comprehensive lexical and morphological analysis of Sinhala language. Section 6 and 7 give details about available lexical resources which we use in this research. Section 8 and 9 describe POS tagging and the Hidden Markov Model implementation algorithm. Section 10 and 11 discuss the Evaluation, testing and the result, and section 12 concludes the paper and describes the future work.

## 2. PREVIOUS WORK ON NLP

Natural language processing history started from Shanon (1948), Kleen (1951) then Chomsky (1956) to Harris (1959), they contributed a lot in early 1950s to formulate the basic concepts and principles of language processing. In the last 50 years of research in language processing, various kinds of knowledge had been captured through the use of small number of formal models or theories. Most of these models and theories are all extracted from the standard toolkit of Computer Science, Mathematics and Linguistics. Among the most important elements in these toolkits are state machine, formal rules system, logic as well as probability theory and other machine learning tools [7]. But in the last decade, probabilistic and data-driven models had become quite standard throughout the natural language processing.

For English, there are many POS taggers available: employing machine learning techniques (based on Hidden Markov Models [15]), transformation based error driven learning [10], decision trees [9] and maximum entropy methods [6]. There are some taggers which are hybrid using both stochastic and rule-based approaches. Most of the POS taggers have reached a success, between 92-97 % accuracy. However, these accuracies are aided by the availability of large annotated corpus for English. Further there are few Tagging systems available for South Asian languages like Hindi, Tamil and Bengali [8, 12, 13, 14]. In 2006, a POS tagger was proposed for Hindi, which uses an annotated corpus of 15,562 words and a decision tree based learning algorithm. They reached an accuracy of 93.45% with a tag set of 23 POS tags [14]. For Bengali, a tagger was developed using a corpus based semi-supervised learning algorithm based on HMMs [13].

## 3. PREVIOUS WORK ON SINHALA NLP ANALYSIS

There were some important language analysis work has done for Sinhala language, and created a Tag set [16] and a corpus of one million words [17], which was an important initiative, that gives a substantial influence to perform NLP research on Sinhala language. But unfortunately, the progress of computational linguistic analysis on Sinhala language is far behind than other languages. According to our knowledge, there is no well-known automated POS tagging system available for Sinhala language.

## 4. MORPHOLOGY IN SINHALA LANGUAGE

Sinhala is morphologically rich and agglutinative language, in which root words are inflected in different contexts. In Sinhala, words are defined as written stream of letters forming a sensible understanding to a person that denotes or relation to the physical world or to an abstract concept.

Basic building blocks of Sinhala words are also sound units not the letters, same as English language, which distinguish two broad classes of morphemes: lemma and affixes . The lemma (stem) is the "main" morpheme of the word, supplying the main meaning, while the affixes add "additional" meaning of various kinds. Often Sinhala words are postpositionally inflected with various grammatical features. Sinhala verb inflects to specifying almost everything like gender, singularity or plurality, person markings and represents the tense. Sinhala nouns inflect and specifying singularity or plurality, gender, person marking and case of the noun [18].

According to tradition, below are four main types of words exist in Sinhala language [4, 5]:

1. *Noun - නාම පද.*
2. *Verb - ක්‍රියා පද.*
3. *Upasarga – උපසර්ග පද (no direct matching with English grammar)*
4. *Nipatha – නිපාත පද (no direct matching with English grammar)*

## 5. SINHALA WORD CLASSES

Traditionally the definition of POS has been based on morphological and syntactic functions [7]. Similar to most of other languages, POS in Sinhala language also can be divided into two broad categories: closed class type and open class type. Closed classes are those that have relatively fixed membership. Closed class words are generally function words: which tend to be very short, occur frequently, and play an important role in grammar. By contrast open class is the type that lager numbers of words are belongs in any language, and new words are continually coined or borrowed from other languages. The words that are usually containing main content of a sentence are belonged to open word class category.

In Sinhala, all Nouns and Verbs can be categorized under open word class. But Nipatha and Upasarga behave differently in Sinhala grammar. Words belong to Nipatha and Upasarga are not changed according to time and gender, Upasarga always join with nouns and provide additional (improve) meaning to the noun, therefore, Upasarga are not categorized under any of word classes, but Nipatha can be categorized as closed class words based on their existence.

In-addition to that, Sinhala Pronouns also can be classified as open class words, based on their morphological properties, but also Pronouns can be classified as closed class words, based on their existence of fixed membership in the language. Sinhala Pronouns are forms of noun commonly referring to person, place or things [11].

## 6. POS TAG SET FOR SINHALA LANGUAGE

In Table I, presents the Tag set defined for Sinhala language, which was developed by UCSC under PAN Localization project in 2005 [16], and this tag set contains 26 tags which are mostly based on morphological and syntactical features of Sinhala language. Currently this is the only tag set available for Sinhala Language, and we use this tag set in our research.

However, there are few issues that the authors have encountered during the process of defining the tag set, based on the syntactical complexity of Sinhala Language [16]:

1. *Separation of Participle[1] and Post-positions[2].*
2. *Separation of Compound Nouns - Combination of multiple nouns act as a single noun.*
3. *Multiword - Certain word combination/phrases can function as one grammatical category.*

Table 1 Sinhala Tag Set

|    | Tag   | Description            |
|----|-------|------------------------|
| 1  | NNR   | Common Noun Root       |
| 2  | NNM   | Common Noun Masculine  |
| 3  | NNF   | Common Noun Feminine   |
| 4  | NNN   | Common Noun Neuter     |
| 5  | NNPA  | Proper Noun Animate    |
| 6  | NNPI  | Proper Noun Inanimate  |
| 7  | PRPM  | Pronoun Masculine      |
| 8  | PRPF  | Pronoun Feminine       |
| 9  | PRPN  | Pronoun Neuter         |
| 10 | PRPC  | Pronoun Common         |
| 11 | QFNUM | Number Quantifier      |
| 12 | DET   | Determiner             |
| 13 | JJ    | Adjective              |
| 14 | RB    | Adverb                 |
| 15 | RP    | Particle               |

---

[1] Particle is a word that resembles a preposition.

[2] By definition a post-position follows a noun or a noun phrase.

|    | Tag  | Description            |
|----|------|-----------------------|
| 16 | VFM  | Verb Finite Main      |
| 17 | VNF  | Verb Non Finite       |
| 18 | VP   | Verb Ptharticiple     |
| 19 | VNN  | Verbal Non Finite Noun|
| 20 | POST | Postpositions         |
| 21 | CC   | Conjunctions          |
| 22 | NVB  | Noun in Kriya Mula    |
| 23 | JVB  | Adjective in Kriya Mula|
| 24 | UH   | Interjection          |
| 25 | FRW  | Foreign Word          |
| 26 | SYM  | Not Classified        |

# 7. SINHALA TEXT CORPUS

Corpus is also an important lexical resource in the field of NLP. In this research we use the Beta version of the Corpus developed by the UCSC under PAN Localization project in 2005 [17], which contains around 650 000 words and out of which 70000 distinct words, that comprise of data drawn from different kinds of Sinhala newspaper articles.

# 8. POS TAGGING

Part-of-speech tagging is the process of assigning a part-of-speech or other lexical class marker to each word in a sentence [7]. The input to a tagging algorithm is a string of words and a tag set. The output is a single best tag for each word. For example, here is a sample sentence from Sinhala Text Corpus of a news report about "Silsamadana on a Wesak poya day" which each word tagged with mapping tag using the tag set defined in Table I.

Example: වෙසක්_NNPI පෝය_NNN නිමිත්තෙන්_POST මැයි_NNPI 2_NUM වැනි_QFNUM දා_NNN පැවැති_VP ශීල_JJ වියාපාරයට_NNN ද_RP දෙසියක්_QFNUM පමණ_RP පිරිසක්_NNM සහභාගි_NVB වුහ_VFM ._. [Refer the Sinhala glossary for meaning of Sinhala words]

Sinhala is a morphologically rich and agglutinative language, which words are made up of lexical roots combined with affixes or prefixes. So automatically assigning a tag to each word in a language like Sinhala is very complex. The main challenge in Sinhala POS tagging is solving the complexity of words. Ambiguity is also adding some complexity in the process of tagging, but fortunately most words in Sinhala are unambiguous. The example below shows how a word can be ambiguous in Sinhala language.

Ambiguity is the existence of more than one possible usage of POS in different context. The noun ඉබ්බා (ibba) containing two meanings tortoise and padlock bear intimateness and inanimateness according to the context. The problem of POS tagging is to resolve these ambiguities, choosing the proper tag for the context, not for the word. So this can be resolved by looking at the associated words with the word.

Example:

ඉබ්බො_NNPA දෙන්නෙක් ඉන්නවා (There are two tortoises).

ඉබ්බො_NNPI දෙන්නෙක් දාලා තියනවා (Two padlocks have been fixed).

Most of tagging algorithms fall into one of the two classes: rule-based taggers and stochastic taggers [7]. Rule-based taggers generally involve a large database of hand written disambiguation rules. Stochastic tagger generally resolves tagging ambiguities by using a training corpus to compute the probability of a given word having a given tag in a given context. In addition to that, there are taggers, which use a hybrid approach, which employees both of the above methods to resolve the tagging ambiguity, which is called transformation-based taggers or Brill taggers [7]. Under this research we have tried out the applicability of Stochastic based tagging approach for Sinhala language.

# 9. OUR APPROACH

We next describe the approach and the overall application architecture defined for Sinhala POS tagger in this Research. To find a suitable tagging approach for Sinhala language, we analysed multiple approaches that has already been discussed for other morphologically rich languages and decided to use a well-known stochastic based approach which is known as Hidden Markov Model that has proven evidence of better results for other languages. Probability is the basic principle behind HMM. The model described here follows the concepts given in the reference [7].

The intuition behind all stochastic taggers is simple generalization of the "pick the most-likely tag for this word". For a given sentence or a word sequence, HMM tagger chooses the tag sequence that maximizes:

$$P\ (word\ |\ tag)\ *\ \ P\ (tag\ |\ previous\ n\ tags).$$

HMM tagger generally chooses a tag sequence for a given sentence rather than for a single word. This approach assumes that we are trying to compute the most probable tag sequence $(\hat{T})$ of tags $T = (t_1, t_2, \ldots, t_n)$ for a given sequence of words in the sentence $W = (w_1, w_2, \ldots, w_n)$:

$$\hat{T} = \underset{t\,\epsilon\,T}{argmax}\ P(T|W)$$

where,

$$\underset{t\,\epsilon\,T}{argmax}\ P(T|W)$$

is the set of values of t for which $P(T|W)$ attains its maximum value.
By Bayes law, $P\ (T|W)$ can be expressed as

$$P(T\ |\ W) = \frac{P(T)P(W|T)}{P(W)}$$

So we choose the sequence of tags that gives

$$\hat{T} = \underset{t\,\epsilon\,T}{argmax}\ \frac{P(T)P(W|T)}{P(W)},$$

where,

$$\underset{t\,\varepsilon\,T}{argmax}\;\frac{P(T)\,P(W|T)}{P(W)}$$

is the set of values of *t* for which $(P(T)P(W|T)/P(W))$ attains its maximum value.

Since we are looking for the most likely tag sequence for a sentence given a particular word sequence, the probability of the word sequence $P$ ($W$) will be same for each tag sequence and we can ignore it. So we get

$$\hat{T} = \underset{t\,\varepsilon\,T}{argmax}\;P(T)\,P(W|T)$$

where, $P(T)$ is the Prior probability and $P(W|T)$ is the Likelihood probability.

From the chain rule of probability, we get

$$P(T)P(W|T) = \prod_{i=1}^{n} \begin{array}{l} P(w_i|w_1 t_1 \dots w_{i-1}t_{i-1}t_i)* \\ P(t_i|w_1 t_1 \dots w_{i-1}t_{i-1}) \end{array} \qquad (1)$$

But for a long sequence of words, calculating probabilities like $P$ ($w_i|w_1t_1\dots w_{i-1}t_{i-1}$)*$P$ ($t_i|w_1t_1\dots w_{i-1}t_{i-1}$) is not an easy task, there is no easy way to calculate probability for selecting tag to a word given a long sequence of preceding words. We could solve this problem by making useful simplification: we approximate the probability of a word given all previous words. The probability of the word given the single previous word called bigram model. Bigram model approximates the probability of a word given all the previous words by the conditional probability of the preceding word.

This assumption that the probability of a word depends only on the previous words is called Markov assumption. Markov models are the class of probabilistic models that assume that we can predict the probability of some future unit without looking too far into the past. We can generalize the bigram to the trigram which looks two words into the past [7].

In practice, trigram model is always used in NLP applications. So that let us define the simplifying assumptions for this scenario.

First make the assumption that the probability of a word depends only on its tag, i.e.,

$$P(w_i|w_1 t_1 \dots w_{i-1}t_{i-1}t_i) = P(w_i|t_i) \qquad (2)$$

Next, we make the assumption that the tag history can be approximated by the most recent two tags

$$P(t_i|w_1 t_1 \dots w_{i-1}t_{i-1}) = P(t_i|t_{i-2}t_{i-1}) \qquad (3)$$

From (1), (2) and (3), we get

$$P(T)P(W|T) = P(t_1)P(t_2|t_1)\prod_{i=3}^{n} P(t_i|t_{i-2}t_{i-1})\,[\prod_{i=1}^{n} P(w_i|t_i.$$

Thus, the best tag sequence can be choose, so that it maximize

$$P(t_1)P(t_2|t_1)\prod_{i=3}^{n} P(t_i|t_{i-2}t_{i-1})\,[\prod_{i=1}^{n} P(w_i|t_i)]$$

Now, as usual we can use maximum likelihood estimation from relative frequency to compute these probabilities. We use corpus to find counts of tag sequences of tags $t_{i-2}, t_{i-1}, t_i$ and tags $t_{i-2}, t_{i-1}$, where $t_i$ is the tag $i$ and $t_{i-1}, t_{i-2}$ are previous two tags, and count of $w_i t_i$, where $w_i$ is the word $i$ and $t_i$ is the tag assigned to word $i$.

We compute the probabilities

$$P(t_i | t_{i-2} t_{i-1}) = \frac{c(t_{i-2} t_{i-1} t_i)}{c(t_{i-2} t_{i-1})}$$

and

$$P(w_i | t_i) = \frac{c(w_i t_i)}{c(t_i)}$$

for all $w_i$, where $1 \leq i \leq n$.

## 9.1 The Algorithm

The algorithm explains below is based on the Viterbi Algorithm [7], which is widely used in the NLP applications, that allows considering all the words in the given sentence simultaneously and computes the most likely tag sequence. More formally, the algorithm searches for the best tag sequence for given an observation sequence $W = (w_1, w_2, \ldots, w_n)$ based on the text corpus. Each cell *viterbi*[*t,i*] (a two dimension array with *i*\**j* elements) of the matrix contains the probability of the path which contains the probability for the first *t* observations ends in state *i*. This is the most-probable path out of all possible sequence of the tags of length *t*-1.

The algorithm sets up a probability matrix, with one column for each observation index (*t*) and one row for each state (*i*) in the state graph. The algorithm first creates *t*+2 columns. The first column is the initial observation, which is the start of the sequence, then next corresponds to the first observation, and so on. Then begin with the first column by setting the probability of the start to 1, and other probabilities to 0. For each column of the matrix, that is, for each time index *t*, each cell *viterbi*[*t,j*] will contain the probability of the most likely path to end in that cell *j*. We calculate this probability recursively, by maximizing over the probability of the coming from all possible preceding states. Then we move to the next state; for each of the state *i*, *viterbi*[0,*i*] in column 0, then compute the probability of moving into each of the cell *j*  *viterbi*[1,*j*] in column 1, and finally, the probability for the best path will appear in the final column. Finally back tracing can be done to find the path that gives the best possible tag sequence.

## 9.2 Overall Application Architecture and the Design

Figure 1 shows the overall architecture of the proposed tagger, which is a two-step process that first runs through the tagged corpus and extract the linguistic knowledge. Then it runs through the row text inputs and generating the best tag sequence for the sequence of input words based on the knowledge that gathered from the corpus.

**Lexical Parser:** Checks boundary conditions of each sentences and words as defined in the lexical rules, and prepare for Tokenizing and Pre-processing.
**Tokenization:** Run through the tagged corpus, separate out the words and tags, prepare for probability calculation.

**Probability Calculation**: Calculate the Transition probability and the observation likelihood probability for each pairs of Words, Tag sequences in the corpus as explained in section 9.3 below.

**Viterbi Matrix Analyzer:** Prepare a state graph that has all possible state transitions for the given text input, calculate and assign state transition probability for each transition in the matrix, as explained in section A.

**Tag Sequence Analyzer:** Back trace the viterbi matrix, analyse the maximum probability path and assign tags to each word in the sentence based on highest probability.

## 9.3 Train the Tagger

The next important step is training the tagger. The training method we describe here is based on supervised learning approach. It runs on the corpus, makes use of tagged data and estimates the probabilities of transition, $P(tag \mid previous\ tag)$ and observation likelihood $P(word \mid tag)$ for the HMM.

Then the transition probability $P(t_i|t_{i-1})$ is calculated simply by using the following formula.

$$P(t_i|t_{i-1}) = \frac{c(t_{i-1}t_i)}{Total\ number\ of\ bigrams\ starts\ with\ t_{i-1}},$$

where $c(t_{i-1}t_i)$ is the count of tag sequence $t_{i-1},t_i$ in the corpus.

For calculating observation likelihood probability $P(w_i|t_i)$, we calculate the unigram (unigram model uses only one piece of information, which is the one that is considering) of a word along with its tag assigned in the tagged data. The likelihood probability is calculated simply by the following formula.

$$P(w_i|t_i) = \frac{c(t_iw_i)}{Total\ number\ of\ bigrams\ starts\ with\ t_i},$$

where $c(t_iw_i)$ is the count of word $i$ ($w_i$) is assigned tag $i$ ($t_i$) in the corpus.

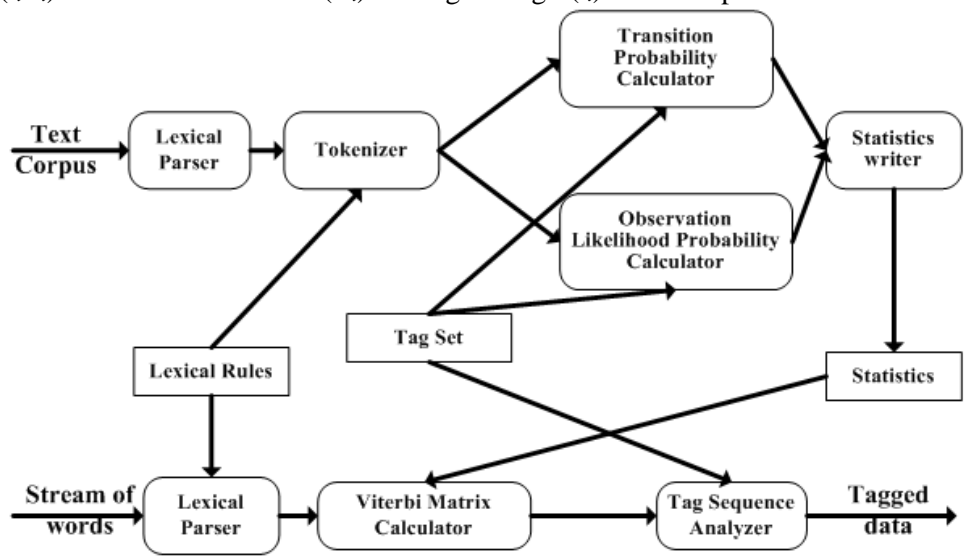

Figure 1. The Architecture of the Tagger

## 10. EVALUATION

The evaluation of the system was mainly driven by train the system using Sinhala text corpus, which comprised of 2754 sentences and 90551 words. The training data set was selected from the Sinhala text corpus developed by UCSC and we used only articles which drown from various Sinhala newspapers. Example given below shows a part of the corpus, in which each word is annotated with corresponding mapping tag.

Example: ඇරිස්ටයිඩ්_NNPA ඔය_PRP පාලනයේ_NNN අවසන්_JVB කාලයේ_NNN දී_VNF කියා_VNF සිටි_VP පරිදි_POST ,_, ඔහුගේ_PRP ජනාධිපතිත්වයේ_NNN ඉතිරි_NVB කාල_NNN සීමාව_NNN තුළ_POST විරුද්ධවාදීන්_NNM සමහ_CC බලය_NNN බෙදා_VNF ගැනීමේ_VNN පදනමක්_NNN මත_POST ,_, අර්බුදය_NNN මේ_DET තරමට_NNN මෝරා_VNF ඒමෙන්_VNN වළකා_JVB ගත_VP හැකිව_? නිබිණි_VFM ._.

The testing was performed on a test data extracted from the corpus, and accuracy was calculated using number of correct tags proposed by the system and total number of words in the sentence/s, by the following formula.

$$Accuracy = \frac{No.\,of\,correct\,tags}{Total\,no.\,of\,words} * 100\%$$

The results were obtained by performing a cross validation over the corpus. The accuracy for known and unknown words was also measured separately.

## 11. RESULT AND DISCUSSION

Testing was done under two classifications: first, tested only with known words (which is already tagged and the tagger is trained), that gives a very high accuracy close to 95%, secondly tested the data set with few unknown words and that gives a less accuracy. The tagger doesn't perform after reaching an unknown word.
Table 2 contains part of test results that were obtained by performing tests for evaluating known word scenarios. Actual and predicted tag assignment for each word in the sentences is shown in the table.

Table 3 below presents the confusion matrix, which summarized the test results given in Table 2. In this confusion matrix, all correct predictions are located in the diagonal of the table. Only one tag assignment has deviated from the actual out of 9 actual NNN tag assignments, system has predicted NNN tags for 7 words, NVB tag was assigned for other two words. In this case, the accuracy of the system has reached to 90.91% for known words scenarios.

Hence, increasing the size of the training corpus is required to increase the tagging accuracy. Not only that, it is required to include data from a wide range of domains that makes the corpus more unbiased and representative, and also further research are required in increasing and optimizing the tagging accuracy for   known words scenarios.

Further, tagging data with unknown words is also an essential need to handle in the tagger. When the system reach an unknown word, current tagger fails to propose a tag, since the system is not trained for that word and the tagging algorithm doesn't have enough intelligence to propose tags for untrained words. So improvements can be suggested to the algorithm by extracting knowledge mainly from open class word category, since new words are coined or browed from other languages more commonly belongs to open word class. Due to fixed number of membership of closed class word category, we can assume that the words belongs to closed class category are well defined in Sinhala grammar and that is fixed. So improvements of the algorithm can be suggested to focus more on words belongs to sub categories of open class words, such as noun, verbs and pronouns. This could be done by incurring some intelligence to the tagger by set of hand written disambiguation rules, and follow the hybrid approach in the tagging algorithm.

Table 2. Test Data

| Test Data | |
|---|---|
| **1** | Predicted: ලාංකිකයින්ට _NNM ආයාචනා _NNN !_. <br><br> Actual: ලාංකිකයින්ට _NNM ආයාචනා _NNN !_. |
| **2** | Predicted: බ්‍රිතාන්‍ය _NNPI ජාතිකයන් _NNM පස් _QFNUM දෙනා _NNM නිදහස් _NVB !_. <br><br> Actual: බ්‍රිතාන්‍ය _NNPI ජාතිකයන් _NNM පස් _QFNUM දෙනා _NNM නිදහස් _NNN !_. |
| **3** | Predicted: නිදහස් _NNN සන්ධානයෙන් _NNN ශ්‍රී _NNPI ලාංකිකයින්ට _NNM ආයාචනා _NVB !_. <br><br> Actual: නිදහස් _NNN සන්ධානයෙන් _NNN ශ්‍රී _NNPI ලාංකිකයින්ට _NNM ආයාචනා _NNN !_. |
| **4** | Predicted: බ්‍රිතාන්‍ය _NNPI මහ _JJ කොමසාරිස් _NNM උතුරට _NNN යයි _VFM ._. <br><br> Actual: බ්‍රිතාන්‍ය _NNPI මහ _JJ කොමසාරිස් _NNM උතුරට _NNN යයි _VFM ._. |
| **5** | Predicted: දිග් _JJ විජයෙන් _NNN ධර්ම _NNN විජය _NNN කරන _VP ඇල්ගන් _NNPI තලේබාන්වරු _NNPA ._. <br><br> Actual: දිග් _JJ විජයෙන් _NNN ධර්ම _NNN විජය _NNN කරන _VP ඇල්ගන් _NNPI තලේබාන්වරු _NNPA ._. |

Further, our research opens more areas to continue researches on tagging Sinhala language, which leads more work to be carried out on finding optimization techniques and unknown word handling approaches.

Table 3. Confusion Matrix of the Test Result

| | | Predicted | | | | | | |
|---|---|---|---|---|---|---|---|---|
| | | *NNM* | *NNN* | *NNPI* | *NVB* | *JJ* | *VFM* | *VP* |
| **Actual** | *NNM* | 5 | | | | | | |
| | *NNN* | | 7 | | 2 | | | |
| | *NNPI* | | | 4 | | | | |
| | *NVB* | | | | 0 | | | |
| | *JJ* | | | | | 2 | | |
| | *VFM* | | | | | | 1 | |
| | *VP* | | | | | | | 1 |

## 12. CONCLUSION AND FUTURE WORK

In this research, our effort was mainly focused on giving a push to NLP and computational linguistics analysis for Sinhala language by developing a tagging system (according to our knowledge, there is no language specific tagging system available for Sinhala language). In this paper, we have described the POS tagging approach that we have developed, which is an implementation of stochastic model approach based on HMM. An algorithm has been produced for the said model. The model was tested against 90551 words, 2754 sentences of Sinhala text corpus, the tagger gave more than 90% accuracy for known words, but the system is not performing well for the text with unknown words yet. So unknown words scenarios are still an open area for further researches.

Though this research produced a tagger for Sinhala language, more research is required in this to improve and optimize the algorithm. Hence, several interesting directions are suggested here for future work.

- Since new words are continuously coming into the language, handling the unknown words (Out-Of-Vocabulary) is required.
- In-addition to disambiguation, there are few other complex scenarios exist in Sinhala language, which separate particles and post-positions, separation of compound nouns, multiword (combination/ phrases can be function as one grammatical category) and separation of using Nipatha (නිපාත) in different contexts, which are not handled in this research.
- Smoothing technique can be applied to get a better outcome.

## ACKNOWLEDGEMENTS

I express my immense gratitude and many thanks to Mr. Harsha Kumara at University of Kelaniya for his invaluable support in providing an initiative to NLP in Sinhala language. Many thanks to Mrs. Kumudu Gamage at the Department of Linguistics, University of Kelaniya for her kind support.

Glossary of Sinhala Terms

|   | Sinhala Term | English Translation |
|---|---|---|
| 1 | වෙසක් පෝය | Vesak Poya |
| 2 | නිමිත්තෙන් | due to |
| 3 | මැයි වැනි 2 දා | on May 2nd |
| 4 | පැවැති | held |
| 5 | ශීල වාහාපාරයට ද | program of observing Sill |
| 6 | දෙසියක් පමණ | around two hundred |
| 7 | පිරිසක් | persons |

|  | **Sinhala Term** | **English Translation** |
|---|---|---|
| 8 | සහභාගි වූහ | participated |
| 9 | ඉබ්බො | tortoises, padlock |
| 10 | දෙන්නෙක් | two |
| 11 | ඉන්නවා | there are |
| 12 | දාලා තියනවා | have been fixed |
| 13 | ඇරිස්ටඩ් | name, Aristed |
| 14 | සිය පාලනයේ | in his rule |
| 15 | අවසන් කාලයේ දි | during the last period of |
| 16 | කියා සිටි පරිදි | as told |
| 17 | ඔහුගේ | his |
| 18 | ජනාධිපතින්වයේ | presidency |
| 19 | ඉතිරි කාල සීමාව තුළ | during remaining time period |
| 20 | විරුද්ධවාදින් | opposition |
| 21 | සමහ | with |
| 22 | බලය | power |
| 23 | බෙදා ගැනීමේ | distribution of |
| 24 | පදනමක් | base |
| 25 | මත | on |
| 26 | අර්බුදය | trouble |
| 27 | මේ තරමට මෝරා ඒමෙන් | Expanded in to this level |
| 28 | වළකා ගත හැකිව තිබිණි | could be avoided |
| 29 | ලාංකිකයින්ට | for Sri Lankans |
| 30 | ආයාචනා | summoned, called |
| 31 | බ්‍රිතාන්‍ය | British |
| 32 | ජාතිකයන් | nationalist |
| 33 | පස් දෙනා | 5 (5 people) |

| | Sinhala Term | English Translation |
|---|---|---|
| 34 | නිදහස් | freedom, is released |
| 35 | සන්ධානයෙන් | united party |
| 36 | මහ කොමසාරිස් | high commissioner |
| 37 | උතුරට | to the north |
| 38 | යයි | went |
| 39 | දිග් විජයෙන් ධර්ම විජය කරන | ruled by religion |
| 40 | ඇල්ගන් තලේබාන්වරු | Afghan  Talabanish |
| 41 | සිංහල ව්‍යාකරණය | Sinhala grammar |
| 42 | භාෂාවේ | of the language |
| 43 | නිවැරදි ව්‍යාකරණය විධි | correct grammar in use |

**Authors**

A.J.P.M.P. Jayaweera is graduated from the University of Colombo, Sri Lanka and has completed Masters in computer science from the University of Kelaniaya, Sri Lanka, and currently working on a Master of Philosophy degree in the field of Natural Language Processing and Computational Linguistics at the same university. Professionally he is a professional Software engineer with 11+ years of experience in divers technologies. A proven career records in enterprise application development which involved providing business critical real time application for leading industries. At present, he is working as a software Architect at Virtusa Pvt Ltd, No 752, Dr. Danister De Silva Mawatha, Colombo 09, Sri Lanka.

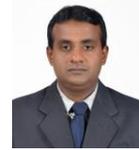

Dr. N. G. J. Dias is graduated from the University of Colombo, Sri Lanka, specializing Mathematics as the main subject. He has completed Masters and Doctoral Degrees in Computer Science from the Queen's University of Belfast, Northern Ireland and University of Wales, College of Cardiff, Cardiff of the United Kingdom respectively. At present, Dr. Dias is a Professor in Computer Science attached to the Department of Statistics & Computer Science of the University of Kelaniya, Sri Lanka. He has been working in the field of Computer Science for the last 30 years and he is the team leader of the Natural Language Processing and Computational Mathematics research groups of the University.

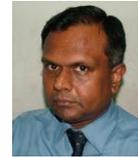